\documentclass[letterpaper, 10 pt, conference]{ieeeconf}  

\IEEEoverridecommandlockouts                              

\usepackage{graphics} 
\usepackage{epsfig} 
\usepackage{times} 
\usepackage{amssymb}  
\usepackage{xcolor}
\usepackage{color, colortbl}
\usepackage{booktabs}
\usepackage{amsmath}

\usepackage{tabularx}
\usepackage{makecell} 
\usepackage{array} 

\usepackage{hyperref}

\usepackage{algorithm}
\usepackage{algpseudocode}
\usepackage{graphicx}
\usepackage[scaled]{helvet} 
\usepackage{booktabs, multirow}
\usepackage{tabularx}
\usepackage[table]{xcolor}
\usepackage{threeparttable}
\usepackage{color}
\usepackage{float}
\usepackage{caption}
\usepackage[justification=raggedright]{caption}
\usepackage[margin=1in]{geometry} 
\usepackage{multicol}
\usepackage{ragged2e}

\definecolor{ref}{RGB}{10,138,218}  
\definecolor{table_color}{RGB}{250,148,147}

\usepackage{cite}
\makeatletter
\let\NAT@parse\undefined
\makeatother
\usepackage{hyperref}
\hypersetup{hypertex=true,
        colorlinks=true,
	linkcolor=ref,
	filecolor=ref,      
	urlcolor=ref,
	citecolor=ref,
}

\title{\LARGE \bf
Every Dataset Counts: Scaling up Monocular 3D Object Detection with Joint Datasets Training
}

\author{Fulong Ma, Xiaoyang Yan, Yuxuan Liu and Ming Liu \\
HKUST(GZ), HKUST \\%
\{fmaaf, xyanaq, yliuhb, eelium\}@connect.ust.hk
\thanks{The authors are with the Robotics and Multi-Perception Laborotary, 
Department of Electronic and Computer Engineering, The Hong Kong University of Science and Technology.
 email:{\tt\small fmaaf@connect.ust.hk yliuhb@connect.ust.hk, eewanglj@ust.hk, eelium@ust.hk}.
}
}

\begin{document}
\maketitle
\thispagestyle{empty}
\pagestyle{empty}


\begin{abstract}

Monocular 3D object detection plays a crucial role in autonomous driving. However, existing monocular 3D detection algorithms depend on 3D labels derived from LiDAR measurements, which are costly to acquire for new datasets and challenging to deploy in novel environments. Specifically, this study investigates the pipeline for training a monocular 3D object detection model on a diverse collection of 3D and 2D datasets. The proposed framework comprises three components: (1) a robust monocular 3D model capable of functioning across various camera settings, (2) a selective-training strategy to accommodate datasets with differing class annotations, and (3) a pseudo 3D training approach using 2D labels to enhance detection performance in scenes containing only 2D labels.
With this framework, we could train models on a joint set of various open 3D/2D datasets to obtain models with significantly stronger generalization capability and enhanced performance on new dataset with only 2D labels.  
We conduct extensive experiments on KITTI/nuScenes/ONCE/Cityscapes/BDD100K datasets to demonstrate the scaling ability of the proposed method. 
Here is our project page: \href{project page}{https://sites.google.com/view/fmaafmono3d}.

\end{abstract}

\section{INTRODUCTION}

\begin{figure*}[t]
    \setlength{\abovecaptionskip}{0pt}
    \setlength{\belowcaptionskip}{0pt}
    \centering
    \includegraphics[width=1.0\linewidth]{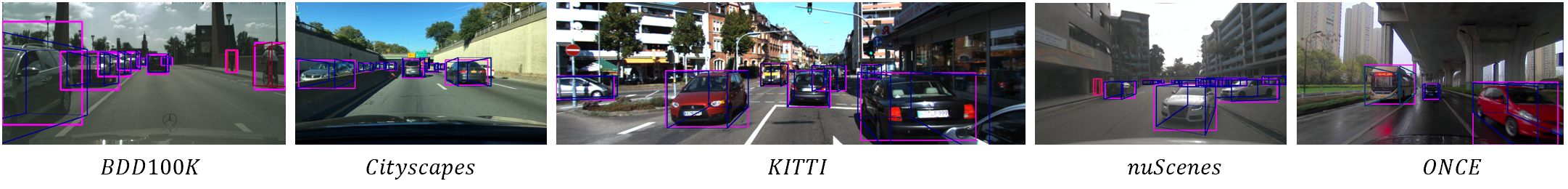}
    \caption{Visualizations of the detection results of our method on five different datasets: BDD100K, Cityscapes, KITTI, nuScenes, and ONCE. 
Our proposed approach allows for algorithm training across multiple datasets with inconsistent camera settings. Furthermore, our method leverages 2D detection labels for training 3D detection algorithms, significantly reducing the annotation costs associated with 3D detection data. This approach enables the model to exhibit robust 3D detection capabilities when applied to new datasets that only provide 2D annotation labels. The pink box represents the 2D detection results.}
    \label{vis_compare}
\end{figure*}

The pursuit of a comprehensive 3D understanding of dynamic environments stands as a cornerstone in the fields of robotics, autonomous driving, and augmented reality. Contemporary 3D detection algorithms leveraging LiDAR point clouds have shown remarkable performance, owing to the enhanced accuracy and density of these point clouds \cite{Yun2018Focal}. Nevertheless, when contrasted with LiDAR, cameras are found to be cost-effective, power-efficient, and offer greater flexibility in installation. This has led to their widespread use in robots and autonomous vehicles, consequently making 3D detection via single cameras an increasingly compelling area of research within robotics and computer vision.

Over the last few years, monocular 3D object detection has experienced a significant evolution. Models employing Bird-Eye-View (BEV) representation have notably amplified potential in settings involving multi-sensor fusion, frequently encountered in autonomous driving scenarios. However, models utilizing front-view representation, the inherent representation of camera images, are not only faster but also simpler to deploy.

Despite these advancements, deploying monocular 3D object detection continues to grapple with major challenges, with data being a principal concern. When aiming to deploy a 3D detection model on a robot equipped with a single camera in a new environment, obtaining 3D data labeled with LiDAR points becomes a hurdle, and fine-tuning the model with data gathered from the robot is unattainable. It is, therefore, paramount that we depend on the pretrained model's generalization capability with almost zero-shot fine-tuning. Furthermore, generating a model with robust generalization performance necessitates expansive datasets, which are considerably costly to compile.

In light of these challenges, we put forth methodologies designed to augment the deployment of monocular 3D detection models through effective utilizations of existing data. Firstly, we explore the intricacies of training vision-based 3D detection models on a combined assortment of public 3D datasets with varied camera parameters. In doing so, we devise an output representation for the monocular 3D detection model that remains invariant to changes in camera parameters and, concurrently, establish a paradigm to train models on datasets with different labels. This approach could substantially boost model scales for individual researchers and developers.

Subsequently, we devise a strategy for training 3D models using 2D data. Several existing methods, such as MonoFlex\cite{MonoFlex}, annotate objects on the heatmap based on the 3D center projected on the image as opposed to the center of 2D bounding boxes. We formulate a training strategy that allows these models to fine-tune with 2D labels. This method enables us to fine-tune existing models with cheaper 2D-labeled data procured from on-site robots, thereby transferring 3D knowledge from public 3D datasets to the target environment. Throughout this paper, we will present experiments conducted to fine-tune a pre-trained model using KITTI 2D data and Cityscapes 2D data as evidence of the potential of our proposed method.

Moreover, by integrating the two proposed methods, we can significantly expand our monocular 3D detection models by training them on a mix of multiple public 3D/2D datasets. This dramatically enlarges the data available during the training phase, thereby significantly enhancing the generalization ability of the resulting model. 
We pre-train a model on a combined set of KITTI, nuScenes, ONCE, Cityscapes, and BDD100K datasets and then fine-tune it on the target dataset using only 2D labels. We subsequently test the model's generalization ability on the target dataset.

The primary contributions of our work can be summarized as:

\begin{itemize}
\item Examination of models such as MonoFlex, and formulation of an output representation robust to varying camera settings, which serves as the foundation for training models on disparate datasets.
\item Development of strategies to train or fine-tune 3D vision models on a blend of 3D/2D datasets.
\item Extensive experiments were conducted on a combined set of KITTI, nuScenes, ONCE, Cityscapes and BDD100K datasets to assess the efficacy of our proposed methods.
\end{itemize}


\section{RELATED WORKS}
\subsection{Monocular 3D Object Detection}

Recent developments in monocular 3D detection can be broadly categorized into two methodologies: bird-eye-view (BEV) methods and perspective-view (PV) methods.

\textbf{Bird-Eye-View Methods:} These methods focus on conducting monocular 3D detection directly within 3D spaces, which simplifies the design of output representation. However, the main challenges arise in the transformation between perspective-view images and features in 3D coordinates. Pseudo-LiDAR represents a series of works that reconstruct 3D RGB-point clouds from monocular depth prediction models, subsequently applying LiDAR-based 3D object detectors to the reconstructed point cloud \cite{Weng2019Plidar,Vianney2019RefinedMPL,Ma2019AM3D,Ku2019MonoPSR}. Another approach directly performs differentiable feature transformation, constructing 3D features from image features, enabling end-to-end training of 3D detection in 3D spaces \cite{Cody2021CaDDN}. These methods often delegate the scale-ambiguity problem to depth prediction sub-networks or attention modules. However, network inferencing in the BEV space heavily relies on 3D labels from LiDAR or direct supervision from LiDAR data, making it challenging to utilize existing 2D datasets or inexpensive 2D labeling tools. Consequently, researchers with limited access to 3D labeled data may encounter difficulties when deploying or fine-tuning networks in new environments.

\textbf{Perspective-View Methods:} These methods, which conduct monocular 3D detection in the original perspective view, are more intuitive. Many one-stage monocular 3D object detection methods are built upon existing 2D object detection frameworks. The main challenge here lies in devising robust and precise encoding/decoding methods to bridge 3D predictions and dense PV features. Various techniques have been proposed, such as SS3D \cite{Jorgensen2019SS3D}, which adds additional 3D regression parameters, and ShiftRCNN \cite{Li2019ShiftRCNN}, which introduces an optimization scheme. Other works like M3DRPN \cite{Brazil2019M3DRPN}, D4LCN \cite{Ding2019D4LCN}, and YoloMono3D \cite{Liu2021YOLOStereo3D} employ statistical priors in anchors to enhance 3D regression accuracy. Additionally, SMOKE \cite{liu2020SMOKE}, RTM3D \cite{Li2020RTM3DRM}, Monopair \cite{Chen2020MonoPair}, and KM3D\cite{KM3D2020Li} utilize heatmap-based keypoints prediction with anchor-free object detection frameworks like CenterNet \cite{zhou2019objects}.

Despite these advancements, most researchers focus on training within a single homogeneous dataset, leading to overfitting in certain camera settings. In this work, we propose a more robust output representation based on MonoFlex \cite{MonoFlex}, enabling network training across different datasets. We also introduce training strategies that allow our model to be trained on 2D datasets as well.

\subsection{Task Incremental Learning}
Task incremental learning represents a specific scenario within incremental learning where models are sequentially trained to recognize new classes across different training phases \cite{riemannian2018IL}. This means that the models are exposed to datasets with varying sets of annotations, particularly when employing a joint set of data labeled with different types. This approach fosters adaptability and flexibility, allowing the model to evolve and accommodate new information without losing its proficiency in previously learned tasks. It represents a significant step towards creating more dynamic and responsive learning systems, capable of adapting to the ever-changing landscape of data and information. Most of the research are focused on Softmax-based multi-class classification \cite{yun2021defense} where activations from different classes all have impacts on each output.

In this paper, we re-formulate the categorical masking strategy in the setting of joint-dataset training under sigmoid-based classification, allowing more flexible dataset usage.

\section{METHODOLOGY}

\subsection{Camera Aware Monoflex Detection Baseline}
Monocular 3D object detection involves estimating the 3D location center $(x, y, z)$, dimensions $(w, h, l)$ and planar orientation $\theta$ of objects of interest with a single camera image. Since most SOTA detectors perform prediction in the camera's front-view, we generally predict the projection of the object center on the image plane $(c_x, c_y)$ instead of the 3D position $(x, y)$. The orientation $\theta$ is further replaced with the observation angle
\begin{equation}
    \alpha = \theta - \text{arctan}(\frac{x}{z}),
\end{equation}
which better conforms with the visual appearance of the object \cite{MonoFlex, Brazil2019M3DRPN}.


 MonoFlex \cite{MonoFlex} is an anchor-free method.
It extracts feature maps from the input images with a DLA \cite{DLA2018Yu} backbone, similar to CenterNet\cite{zhou2019objects} and KM3D \cite{KM3D2020Li}. As an anchor-free algorithm, MonoFlex predicts the center positions of target objects with a heat map.
Monoflex primarily consists of the following components: 2D detection, dimension estimation, orientation estimation, keypoint estimation and depth estimation ensemble.
The depth prediction z is simultaneously estimated from both geometry and direct prediction, and these predictions are adaptively ensembled to obtain the final result.

Since we aim to perform joint training on diverse datasets, where data collection involves different cameras with distinct camera settings, our method needs to overcome the challenges posed by varying camera parameters in order to facilitate effective knowledge transfer across these datasets. Given that MonoFlex exhibits insensitivity to camera parameters, we build upon MonoFlex as the foundation of our approach.

Overall, our approach is similar to MonoFlex, with the main difference being that our method has modifications in the depth prediction component, which takes camera parameters into account. Specifically, in the original MonoFlex paper, the direct regression part of depth prediction assumes an absolute depth of
\begin{equation}
    z_{r} = \frac{1}{\sigma(z_{o})} - 1,
\end{equation}
where $z_{o}$ is the unlimited network output following \cite{Chen2020MonoPair, zhou2019objects} and 
\begin{equation}
\sigma(x) = \frac{1}{1 + e^{-x}}.
\end{equation}
While in our method, we have improved it by taking camera's parameters into account as follows:
\begin{equation}
    z_{r} = (\frac{1}{\sigma(z_{o})} - 1) \times \frac{f_{x}}{f_{x0}} .
\end{equation}
where $f_{x}$ is epresents the focal length of the camera used in the training dataset, while $f_{x0}$ is a hyperparameter, we set its value to 500 in our experiments.


\subsection{Selective Training for Joint 3D Dataset Training}
\label{sec:why-base-vtr-failed}
\begin{figure*}[]
    \setlength{\abovecaptionskip}{0pt}
    \setlength{\belowcaptionskip}{0pt}
    \centering
    \includegraphics[width=1.0\textwidth,height=0.22\textheight]{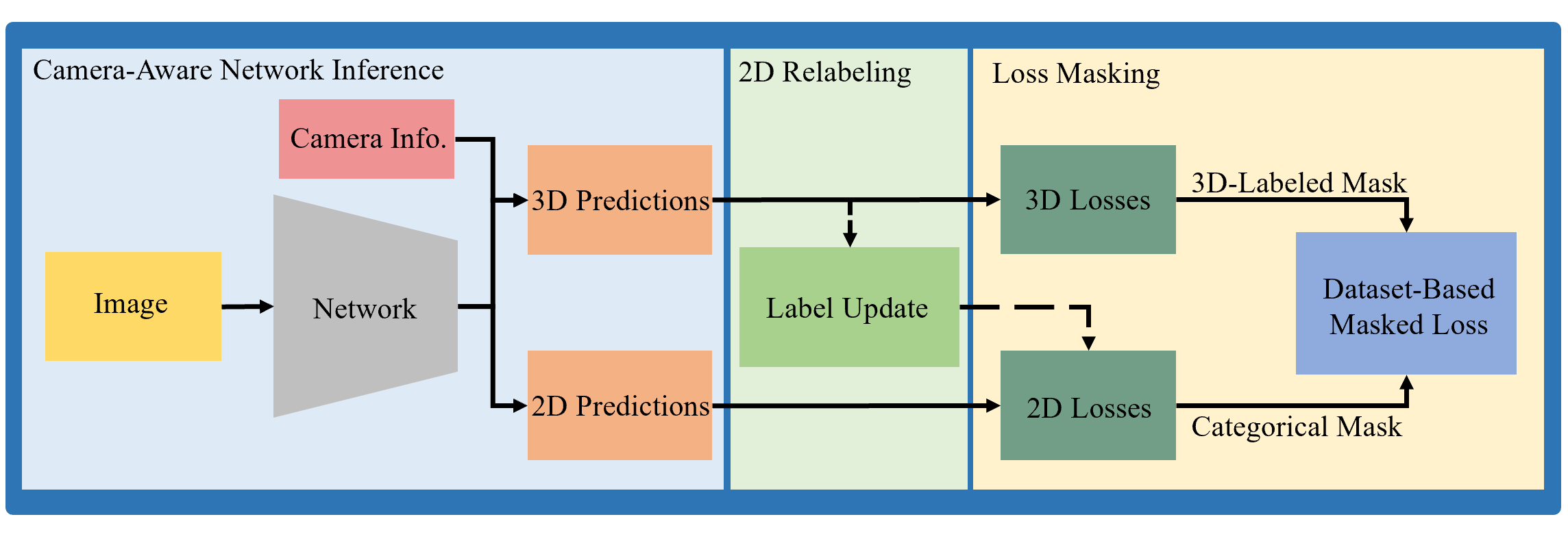}
    \caption{This figure illustrates the training process of our proposed method. It shows how the pre-trained model's inference, combined with the 2D annotations from the dataset, facilitates the training of a 3D detection model on datasets that lack 3D training labels.}
    \label{training}
\end{figure*}

During dataset training, some parts of the dataset are incompletely annotated. For instance, in the case of KITTI, certain categories like "Tram" are not labeled. However, it is not appropriate to treat the data from KITTI as negative samples for the "Tram" category. Therefore, each data point, in addition to its own annotations, is associated with the categories labeled in the respective dataset. This association provides supervision for the network's classification output and is applied only to the categories with annotations. 
Specifically, for different 3D datasets with varying labeled categories, each data frame stores the categories currently labeled in the dataset. When calculating the loss, we do not penalize (suppress) the detection predictions for unannotated categories.
In summary, this approach effectively handles the issue of incomplete annotations during dataset training. It ensures that only annotated categories influence the model's training, and unannotated categories do not lead to erroneous model behavior. The training procedure is shown in Fig. \ref{training}.

\subsection{Regulating 2D Labels of 2D Datasets for Pseudo 3D training}
\begin{figure}[]
    \setlength{\abovecaptionskip}{0pt}
    \setlength{\belowcaptionskip}{0pt}
    \centering
    \includegraphics[width=1.0\linewidth]{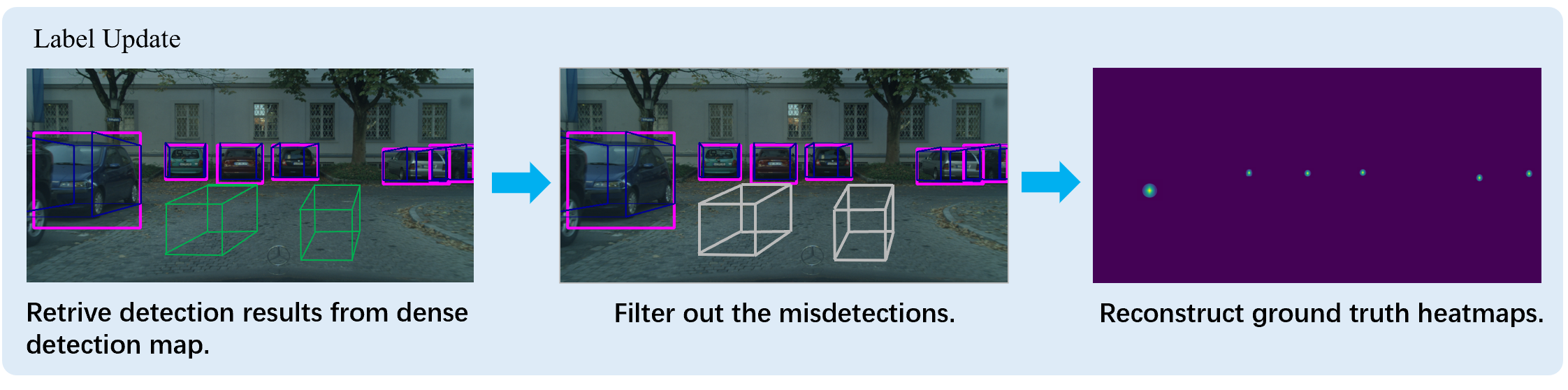}
    \caption{The figure depicts the training label update process. In the left image, the pre-trained 3D detection model's predictions on new data are shown, which may include some erroneous detections, as indicated by the green boxes. The middle image illustrates the process of identifying and filtering out these erroneous detections, marking them in gray based on the matching results. The right image represents the reconstruction of the ground truth heatmap using the pseudo 3D labels.}
    \label{label_update}
\end{figure}

In the Monoflex method, the center of the heatmap is determined by projecting the 3D center. For data with only 2D annotations, we have no direct means of generating supervision signals for the object's 3D center. 
Therefore, we need to find a way to generate 3D detection supervision information from 2D labels. We propose a novel method to train 3D detection algorithms solely relying on 2D detection labels. Specifically, we start by feeding data with only 2D annotations into a pre-trained 3D detection model and set a very low score threshold to enable the model to produce multiple detection results (including both 2D and 3D detection results). This step may include some erroneous or inaccurate detections.
Next, we use the 2D training labels from the new dataset to match them with the 2D detection results obtained in the previous step, filtering out erroneous or inaccurate detections to obtain pseudo 3D training labels. Finally, we reconstruct the ground truth heatmap and 2D detection map (as shown in Fig. \ref{label_update}), and ultimately, we calculate the loss between the model predictions and the pseudo 3D labels only on the heatmap and 2D detection map.
The method of training a monocular 3D model using 2D annotated data is illustrated in Algorithm \ref{alg:post}, we refer to it as \textit{Pseudo 3D Training with 2D Labels}.
\begin{algorithm}
    \caption{Pseudo 3D training with 2D Labels}\label{alg:post}
    \textbf{Input:}  Dense Detection Maps $F$, Labeled 2D boxes $B_{t}$\\
    \textbf{Output:} Loss $l$
    \begin{algorithmic}[1]
    \State {\textbf{Initialization:}}
    \State $B_{p}:$ Detection results from dense detection maps
    \State $B_{t}^{'}:$ Pseudo ground truth
    \State $M:$ IoU matrix
    \State {\textbf{Main Loop:}}
    \State Retrieve detection results $B_{p}$ from dense detection maps $F$.
    \State Compute IoU matrix $M$ between $B_p$ and $B_t$.
    \State Compute the matching with the minimum cost, take 3D centers of $B_p$ as ground truth for $B'_t$.
    \For {each matched $b_p$, $b_t$}
        \If{$cost_i > eps$}
            \State Remove mis-detection box from $B'_t$.
        \EndIf
    \EndFor
    \State Reconstruct ground truth heatmaps and 2D detection maps $F_t$ from $B'_t$.
    \State Compute Loss $l$ with pseudo ground truth $B'_t$ only on heatmaps and 2D detection maps.
    \end{algorithmic}
\end{algorithm}



\begin{figure}[t]
    \setlength{\abovecaptionskip}{0pt}
    \setlength{\belowcaptionskip}{0pt}
    \centering
    \includegraphics[width=1.0\linewidth]{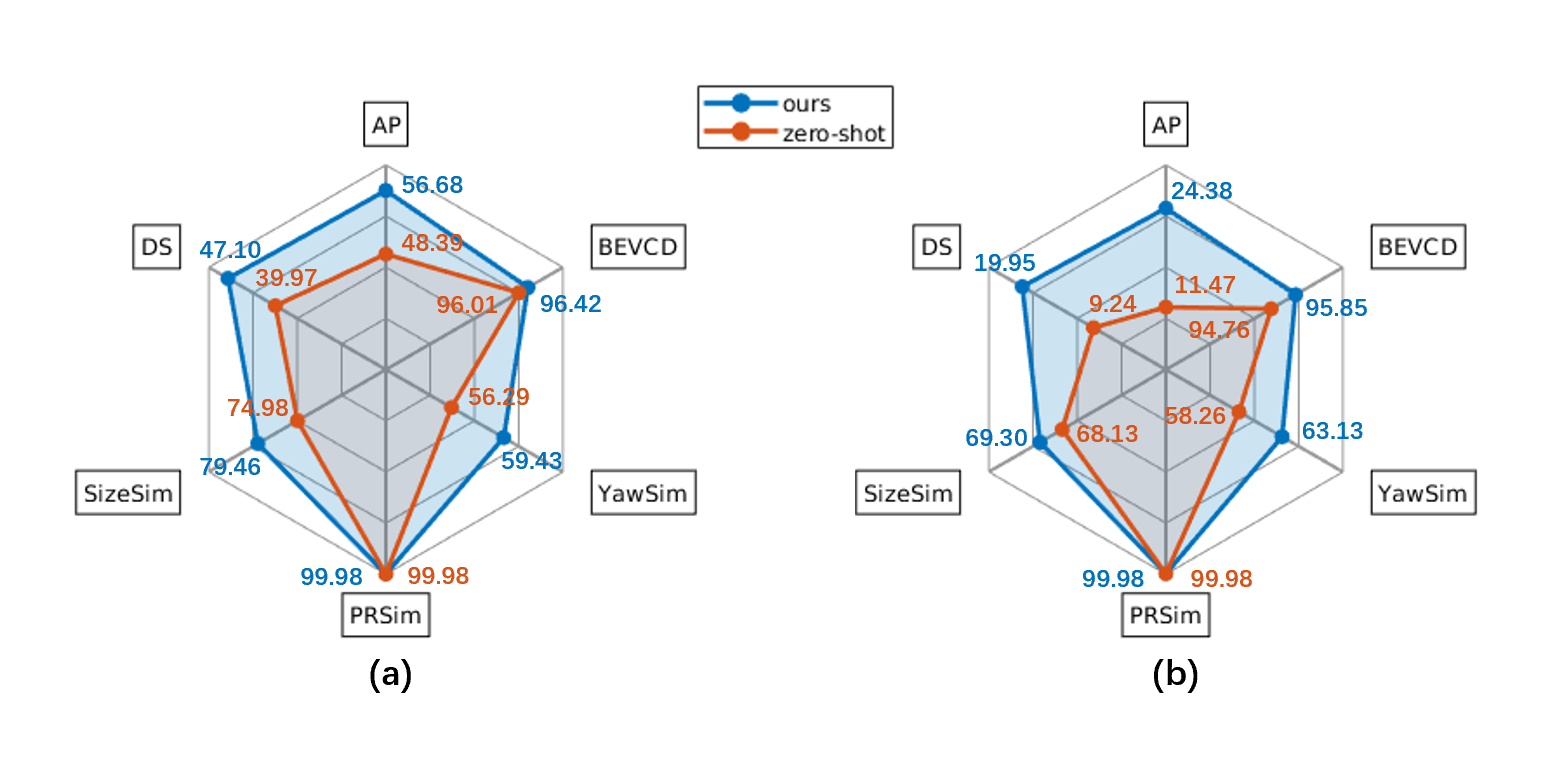}
    \caption{The comparison between our method and zero-shot on the Cityscapes dataset shows that our method outperforms zero-shot in all the evaluation metrics.}
    \label{spider_plot}
\end{figure}

\section{EXPERIMENTS}
XXX


\begin{table*}[h!]
    \centering
    \caption{Detection results on the ``Car'', ``Pedestrian'', and ``Cyclist'' categories on the KITTI dataset. }
    \renewcommand{\arraystretch}{1.1} 
    \setlength\tabcolsep{3pt} 
    \begin{tabular*}{0.96\textwidth}{ c c | c c c | c c c | c c c }
        \cline{1-11}
       \multirow{2}{*}{\bf } & \multirow{2}{*}{\bf Methods} & \multicolumn{3}{c}{\bf Car}   & \multicolumn{3}{c }{\bf Pedestrain} & \multicolumn{3}{c}{\bf Cyclist}  \\ \cline{3-11}
         & & {\bf Easy (\%)} & {\bf Moderate (\%)}& {\bf Hard (\%)}  &{\bf Easy (\%)}&{\bf Moderate(\%)}& {\bf Hard (\%)} &{\bf Easy (\%)}&{\bf Moderate(\%)}& {\bf Hard (\%)}  \\ \cline{1-11}
        \multirow{3}{*}{$2D$} &Zero-shot & 93.94& 81.92& 71.75 &  57.53 & 50.24 & 44.36 & 53.65  & 46.24 & 40.18  \\
        &{ Ours}  & 99.54 & 96.32 
                        & 88.60  &  82.62 
                        & 76.59 & 67.54 & 87.67  & 79.45 & 72.59  \\
        &\cellcolor{table_background_color} { Improvement}  & \cellcolor{table_background_color}+5.60 &\cellcolor{table_background_color}+14.40 
        &\cellcolor{table_background_color}+16.85   
        &\cellcolor{table_background_color}+26.09 
                        &\cellcolor{table_background_color}+26.35  &\cellcolor{table_background_color}+23.18  &\cellcolor{table_background_color}+34.02   &\cellcolor{table_background_color}+33.21 &\cellcolor{table_background_color}+32.41   \\
        \cline{1-11}
        \multirow{6}{*}{$3D$} &Zero-shot (3D) & 30.44 & 23.19  & 20.21  &  6.80 & 5.82 & 5.07 &  0.64 & 0.37 &0.35  \\
        &{ Ours (3D)}  & 48.99  & 33.29  
                        & 28.41    &  14.09 
                        & 12.95 & 10.79 & 1.57  & 0.84 & 0.81  \\
        & \cellcolor{table_background_color} { Improvement}   & \cellcolor{table_background_color}+18.55   &\cellcolor{table_background_color}+10.10
                        &\cellcolor{table_background_color}+8.20   &\cellcolor{table_background_color}+7.29
                        &\cellcolor{table_background_color}+7.13 &\cellcolor{table_background_color}+5.72 &\cellcolor{table_background_color}+0.93   & \cellcolor{table_background_color}+0.47 & \cellcolor{table_background_color}+0.46  \\
        &{Zero-shot (BEV)}  & 38.20  & 28.85  
                        & 25.42    &  7.50 
                        & 6.44 & 5.43 &  1.29 & 0.82 & 0.53  \\
        &{Ours (BEV)}  & 56.87  & 39.17  
                        & 33.79    &  15.40 
                        & 13.72 & 11.37 & 1.86  & 1.06 & 1.01  \\
        &\cellcolor{table_background_color}{ Improvement}  &\cellcolor{table_background_color}+18.67   
        &\cellcolor{table_background_color}+10.32
                        &\cellcolor{table_background_color}+8.37     & \cellcolor{table_background_color}+7.90 
                        & \cellcolor{table_background_color}+7.28 &\cellcolor{table_background_color}+5.94  &\cellcolor{table_background_color}+0.57   & \cellcolor{table_background_color}+0.24 & \cellcolor{table_background_color}+0.48  \\
        \cline{1-11}
    \end{tabular*}
    
    \label{tab:test_results}
\end{table*}

\begin{table*}[h!]
    \centering
    \caption{Object detection results of category ``Car'' on the Cityscapes dataset. }
    \renewcommand{\arraystretch}{1.0} 
    \setlength\tabcolsep{10pt} 
    \begin{tabular*}{0.85\textwidth}{ c c c c c c c c }
        \cline{1-7}
        \multirow{2}{*}{\bf  Methods} & \multicolumn{6}{c }{\bf Metrics}      \\ \cline{2-7}
          & {\bf DS (\%)} & {\bf AP (\%)}& {\bf BEVCD (\%)}  &{\bf YawSim (\%)}&{\bf PRSim(\%)}& {\bf SizeSim (\%)}   \\ \cline{1-7}
      
        Zero-shot & 32.92 & 36.44  & 95.73  & 90.12 & 99.98 & 75.52  \\
        
        Ours  & 56.94  & 61.49  
                        & 96.42     & 92.27  
                        & 99.98 & 81.70  \\
        \cline{1-7}
       \rowcolor{table_background_color} Improvement  & +24.02 & +25.05 & +0.69  & +2.15  &  0 & +6.18  \\
        \cline{1-7}
    \end{tabular*}
    
    \label{tab:test_results_city_car}
\end{table*}

      
        
    

\begin{table*}[h!]
    \centering
    \caption{Object detection results of category ``Truck'' on the Cityscapes dataset. }
    \renewcommand{\arraystretch}{1.0} 
    \setlength\tabcolsep{10pt} 
    \begin{tabular*}{0.87\textwidth}{  c c c c c c c }
        \cline{1-7}
        \multirow{2}{*}{\bf  Methods} & \multicolumn{6}{c}{\bf Metrics}      \\ \cline{2-7}
          & {\bf DS (\%)} & {\bf AP (\%)}& {\bf BEVCD (\%)}  &{\bf YawSim (\%)}&{\bf PRSim(\%)}& {\bf SizeSim (\%)}   \\ \cline{1-7}
        
        Zero-shot & 10.26 & 11.47  & 93.64  &  99.87 & 99.98 & 64.55  \\
        
        Ours  & 23.38  & 25.18  
                        &94.49     &  99.93 
                        & 99.98 & 77.05  \\
        \cline{1-7}
       \rowcolor{table_background_color} Improvement  & +13.12  & +13.71 & +0.85 & +0.06 & 0 & +12.50 \\
        \cline{1-7}
    \end{tabular*}
    
    \label{tab:test_results_city_truck}
\end{table*}

        
        
    

\begin{table*}[h!]
    \centering
    \caption{Object detection results of category ``Bicycle'' on the Cityscapes dataset. }
    \renewcommand{\arraystretch}{1.0} 
    \setlength\tabcolsep{10pt} 
    \begin{tabular*}{0.87\textwidth}{  c c c c c c c }
        \cline{1-7}
        \multirow{2}{*}{\bf  Methods} & \multicolumn{6}{c}{\bf Metrics}      \\ \cline{2-7}
          & {\bf DS (\%)} & {\bf AP (\%)}& {\bf BEVCD (\%)}  &{\bf YawSim (\%)}&{\bf PRSim(\%)}& {\bf SizeSim (\%)}   \\ \cline{1-7}
        
        Zero-shot & 0.02 & 0.03  & 93.14  &  72.42 & 99.98 & 52.91  \\
        
        Ours  & 2.37  & 2.80  
                        &96.64    &  77.63 
                        & 99.98 & 64.65  \\
        \cline{1-7}
        \rowcolor{table_background_color} Improvement  & +2.35& +2.77 & +3.50  & 5.21  &0  & +11.74 \\
        \cline{1-7}
    \end{tabular*}
    
    \label{tab:test_results_city_Bicycle}
\end{table*}

        
        
    

\subsection{Datasets Reviews}

We evaluate the proposed networks on the KITTI 3D object detection benchmark \cite{Geiger2012KITTI} and Cityscapes dataset \cite{gahlert2020cityscapes}. The KITTI dataset consists of 7,481  training frames and 7,518 test frames. Chen \textit{et al.} \cite{Chen2015kittisplit} split the training set into 3,712 training frames and 3,769 validation frames. The Cityscapes dataset contains 5000 images split into 2975 images for training, 500 images for validation, and 1525 images for testing. 




\subsection{Evaluation Metrics}
\subsubsection{KITTI 3D}
All the testing and validation results, are evaluated with 40 recall positions ($AP_{40}$), following Simonelli \textit{et al.} \cite{Simonelli2019MonoDIS} and the KITTI team. Such a protocol is considered to be more stable than the $AP_{11}$ proposed in the Pascal VOC benchmark \cite{Everingham10pascal}.
\subsubsection{Cityscapes 3D}
Following \cite{gahlert2020cityscapes}, we use these five metrics: 2D Average Precision (AP), Center Distance (BEVCD), Yaw Similarity (YawSim), Pitch-Roll Similarity (PRSim), Size Similarity (SizeSim) and Detection Score (DS) to evaluate the performance on Cityscapes 3D dataset. Among them, DS is a combination of the first five metrics and computed as:
\begin{footnotesize}
\begin{equation}
 DS = AP \times \frac{BEVCD+YawSim+PRSim+SizeSim}{4}. 
\end{equation}
\end{footnotesize}
For details, please refer to the paper \cite{gahlert2020cityscapes}.

\subsection{Experiment Setup}
To demonstrate the superiority of our method over zero-shot approaches, when testing on the KITTI dataset, we designed our experiments as follows:

\begin{itemize}
        \item We initially pre-trained our model on four datasets: BDD100K, nuScenes, ONCE, and Cityscapes.
        \item With the pre-trained model, we evaluated its zero-shot detection performance on the KITTI dataset.
        \item Subsequently, we fine-tuned the model using our method, which involves training a 3D detection model using 2D training labels from KITTI.
        \item Finally, we obtained detection results on the KITTI dataset based on our method.
    
\end{itemize}

When testing on the Cityscapes dataset, we followed the same experimental setup on the KITTI dataset.



\subsection{Experiment Results and Comparison}
The quantitative results of the Car category and Pedestrain category on KITTI dataset are reported in Table~\ref{tab:test_results}, while quantitative results of Car category and Truck category on Cityscapes dataset are shown in Table~\ref{tab:test_results_city_car} and Table~\ref{tab:test_results_city_truck} respectively.
We also utilized radar charts to visualize the performance on the Cityscapes dataset, as depicted in Fig. \ref{spider_plot}. Our proposed method exhibits significant improvements over zero-shot learning across five key metrics: AP (Average Precision), BEVCD (Bird's Eye View Center Distance), YawSim (Yaw Similarity), SizeSim (Size Similarity), and DS (Detection
Score). Fig. \ref{vis_compare} illustrates the qualitative results on the Cityscapes dataset.

From Table~\ref{tab:test_results}, we can observe that our method has achieved significant improvements in both 3D and 2D detection tasks compared to zero-shot learning. Specifically, for the 3D detection task, our method has shown an improvement in $AP_{3D}$ for the "Car" category in the easy, moderate, and hard difficulty levels by 16.42\%, 4.27\%, and 0.20\%, respectively. In the "Pedestrian" category, the AP improvements are even more substantial, with increases of 50.7\%, 32.73\%, and 35.34\% for the easy, moderate, and hard difficulty levels, respectively.
The improvements in $AP_{2D}$ for the "Car" category in the easy, moderate, and hard difficulty levels by 5.54\%, 11.25\%, and 12.39\%, respectively, as well as the AP enhancements of 5.69\%, 6.83\%, and 8.80\% in the "Pedestrian" category across the same difficulty levels for the 2D detection task. These improvements in 2D detection performance are easily comprehensible because we trained the algorithm comprehensively on the new dataset using 2D training labels.

Table~\ref{tab:test_results_city_car} and Table~\ref{tab:test_results_city_truck} present the detection results for the "Car" and "Truck" categories on the Cityscapes dataset. Quantitative results indicate that our method, when compared to the zero-shot approach, has shown significant improvements in various metrics, except for the PRSim metric (as we only focused on the yaw angle, considering pitch and roll angles to be 0). Specifically, in the "Car" category, we observed an increase of 17.13\% in AP, a 4.17\% improvement in BEVCD, a 5.58\% improvement in YamSim, a 5.97\% improvement in SizeSim, and a remarkable 17.84\% enhancement in DS. 
In the "Truck" category, AP, BEVCD, YawSim, SizeSim, and DS have improved by 112.55\%, 1.16\%, 8.36\%, 17.17\%, and 115.91\%, respectively.
From these experimental results, we can also observe that our method demonstrates a more significant improvement in detection performance for categories with fewer instances in the dataset. This is due to the fact that the model has less knowledge about categories with fewer instances, leading to weaker generalization on new datasets compared to categories with more instances. After pseudo 3D learning using 2D labels on the new dataset, we achieved a substantial improvement in detection performance.

\begin{figure}[]
    \setlength{\abovecaptionskip}{0pt}
    \setlength{\belowcaptionskip}{0pt}
    \centering
    \includegraphics[width=1.0\linewidth]{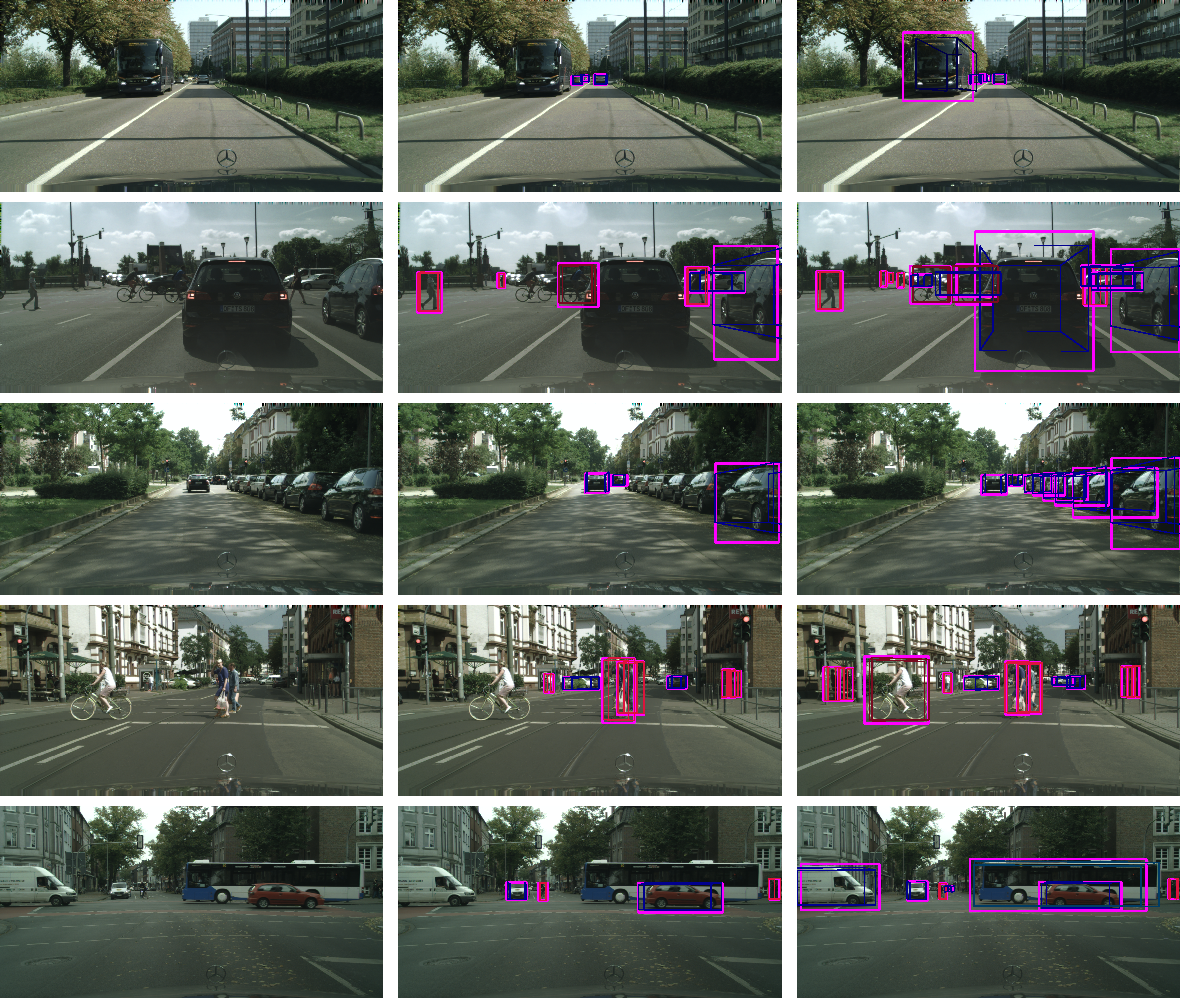}
    \caption{The qualitative results on the Cityscapes dataset. The leftmost column contains the original images, the middle column displays the zero-shot results, and the rightmost column shows the results obtained using our method. The pink boxes represent 2D detection results.}
    \label{vis_compare}
\end{figure}

\section{CONCLUSIONS}
XXX

In this paper, we initially conducted research on models such as MonoFlex and developed strategies that are resilient to changes in camera intrinsics. These strategies allow the models to be trained on diverse datasets. Additionally, we designed a learning approach that enables monocular 3D detection models to acquire 3D detection knowledge based solely on 2D labels, even in datasets that only provide 2D training labels. Lastly, we carried out extensive experiments on a combination of datasets, including KITTI, nuScenes, Cityscapes, and others. The experimental results demonstrated the efficacy of our approach.


Despite its success, our work does have limitations. First, our method is currently applicable only to models that are insensitive to camera parameters, such as MonoFlex. For models where the influence of camera parameters cannot be disregarded.
Moreover, when encountering new categories in a novel dataset, the lack of relevant supervision from previous datasets may result in suboptimal detection performance for these new categories. 
Overcoming these limitations demands continuous optimization and improvement of our approach.


\addtolength{\textheight}{-2cm}   






\clearpage



\bibliographystyle{IEEEtran}
\bibliography{IEEEfull}

\end{document}


\maketitle
\thispagestyle{empty}
\pagestyle{empty}



\section{}




\bibliographystyle{unsrt}

\bibliography{reference}
\balance